\documentclass[conference]{IEEEtran}
\IEEEoverridecommandlockouts
\usepackage{cite}
\usepackage{amsmath,amssymb,amsfonts}
\usepackage{algorithmic}
\usepackage{algorithm}
\usepackage{graphicx}
\usepackage{epstopdf}
\usepackage{textcomp}
\usepackage{xcolor}
\usepackage{svg} 
\usepackage{booktabs}
\usepackage[table]{xcolor}
\usepackage{multirow}
\usepackage{threeparttable}
\usepackage{subcaption}
\usepackage{hyperref}
\usepackage[section]{placeins}
\def\BibTeX{{\rm B\kern-.05em{\sc i\kern-.025em b}\kern-.08em
    T\kern-.1667em\lower.7ex\hbox{E}\kern-.125emX}}
\begin{document}

\title{Adversarial and Score-Based CT Denoising: CycleGAN vs Noise2Score}

\author{\IEEEauthorblockN{Abu Hanif Muhammad Syarubany (20258259)}
\IEEEauthorblockA{
\textit{Korea Advanced Institute of Science \& Technology (KAIST)}\\
Daejeon, South Korea \\
hanif.syarubany@kaist.ac.kr}}

\maketitle

\begin{abstract}
We study CT image denoising in the \emph{unpaired} and \emph{self-supervised} regimes by evaluating two strong, training-data–efficient paradigms: a CycleGAN-based residual translator and a Noise2Score (N2S) score-matching denoiser. Under a common evaluation protocol, a configuration sweep identifies a simple Standard U-Net backbone within CycleGAN (\(\lambda_{\text{cycle}}{=}30\), \(\lambda_{\text{iden}}{=}2\), \(ngf{=}ndf{=}64\)) as the most reliable setting; we then train it to convergence with a longer schedule. The selected CycleGAN improves the noisy input from 34.66\,dB/0.9234 SSIM to 38.913\,dB/0.971 SSIM and attains an estimated score of 1.9441 and an unseen-set (Kaggle leaderboard) score of 1.9343. Noise2Score, while slightly behind in absolute PSNR/SSIM, achieves large gains over very noisy inputs, highlighting its utility when clean pairs are unavailable. Overall, CycleGAN offers the strongest final image quality, whereas Noise2Score provides a robust pair-free alternative with competitive performance. You can access this github repository for more details:  \href{https://github.com/hanifsyarubany/CT-Scan-Image-Denoising-using-CycleGAN-and-Noise2Score}{\textcolor{blue}{\underline{github repository}}}.
\end{abstract}


\section{Introduction}
Deep-learning denoisers have become the de facto approach for low-dose CT (LDCT), but paired low/normal-dose data remain scarce. Unpaired adversarial learning addresses this limitation by translating LDCT to an NDCT-like domain using cycle and identity constraints, while self-supervised score-matching methods denoise directly from noisy images without clean targets. In this work we analyze two representative, high-performing instances of these families: a \textbf{CycleGAN}~\cite{Zhu2017CycleGAN} residual translator with U-Net variants and least-squares GAN (LSGAN) losses, and a \textbf{Noise2Score} denoiser~\cite{Kim2021Noise2Score} trained via an amortized residual DAE~\cite{Lim2020ARDAE} and applied with Tweedie’s estimator at inference. Our goal is to clarify the trade-offs between these pair-free regimes on CT, under a shared evaluation setup. 

We first run a configuration sweep for CycleGAN~\cite{Zhu2017CycleGAN} covering three generator families (Standard U-Net~\cite{Ronneberger2015UNet}, ResU-Net\(+\)~\cite{Gurrola2021ResidualDenseUNet}, Attention U-Net~\cite{Oktay2018AttentionUNet}) and key weights \((\lambda_{\text{cycle}},\lambda_{\text{iden}})\), all trained with Adam (\(\beta_1{=}0.5,\beta_2{=}0.999\)), \(2\times10^{-4}\) learning rate, and 100 epochs, using \(\mathrm{PSNR}/40+\mathrm{SSIM}\) as the estimated score. The best generalizing model is the \emph{Standard U-Net} with \(\lambda_{\text{cycle}}{=}30\), \(\lambda_{\text{iden}}{=}2\), \(ngf{=}ndf{=}64\); we subsequently train this setting to convergence (1000 epochs, \(lr_G{=}10^{-4}\), \(lr_D{=}2\times10^{-4}\), decay at epoch 100) and report both test-set metrics and an external reference on the public Kaggle leaderboard at \href{https://www.kaggle.com/competitions/ai-618-2025-f-midterm-project-ct-denoising/leaderboard}{\textcolor{blue}{\underline{this competition}}}.

\section{Related Work}

Paired-supervised CNNs (e.g., RED-CNN~\cite{Chen2017REDCNN}) initiated deep-learning LDCT denoising but require aligned low-/normal-dose pairs that are scarce in clinical practice. Unpaired adversarial learning addresses this by translating LDCT to an NDCT-like domain without pairs via cycle consistency. Representative systems include CycleGAN-based pipelines~\cite{Tan2022SKFCycleGAN} and selective-kernel/attention–residual variants that better suppress quantum noise while preserving edges and fine anatomy; these consistently report competitive PSNR/SSIM on clinical and Mayo datasets and remain a strong backbone for pair-free denoising~\cite{Li2020UnpairedLDCT,Tan2022SKFCycleGAN}. Recent reviews synthesize these unpaired strategies and highlight recurring risks (texture loss, hallucination) that motivate structural priors and residual designs.

Orthogonally, self-supervised denoisers remove the need for clean targets. Noise2Score (N2S) reframes denoising as posterior mean estimation via Tweedie’s formula and learns the score from only noisy images, yielding a model-grounded, noise-model–agnostic objective that competes with (and often surpasses) prior self-supervised methods~\cite{Kim2021Noise2Score}. Contemporary surveys position cycle-consistency and score-matching as complementary blind/unpaired regimes for LDCT~\cite{Zhao2025Review}. Our method integrates these strands by adopting a CycleGAN-based \emph{residual} translator to protect anatomy, coupled with Noise2Score to provide pair-free regularization/refinement.

\section{Methodology}
\subsection{CycleGAN-Based Residual Learning for CT Denoising}

\begin{figure}[H]
    \centering
    \includegraphics[width=0.99\linewidth]{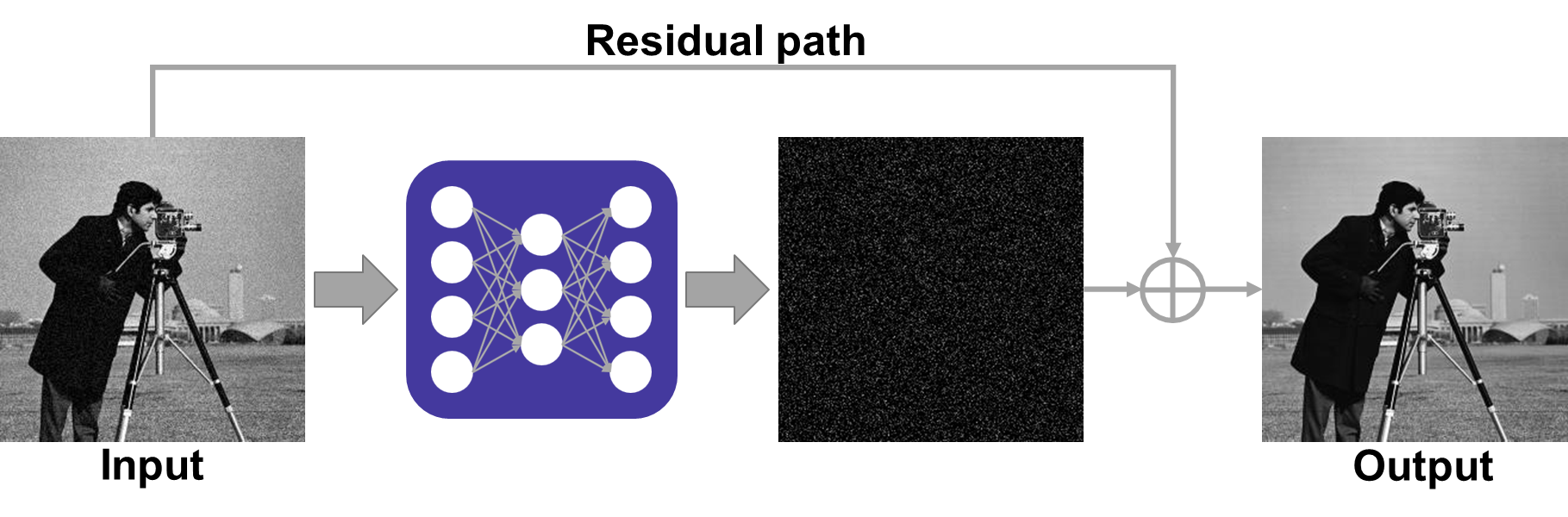} 
    \caption{\textbf{Residual-learning overview for the CycleGAN generator:} the network predicts a noise residual that is added to the input via a global skip to produce the denoised image.}
    \label{fig:residual_overview}
\end{figure}

To address the noise reduction problem in low-dose CT imaging without requiring paired training data, we design a CycleGAN-based residual learning framework as shown in Fig.~\ref{fig:residual_overview}. The model learns two mappings between low-dose and normal-dose domains, while the residual connections ensure that the generator focuses on learning noise-specific corrections rather than reconstructing the entire image content.

\subsubsection{Generator Architecture}

\begin{figure*}[htbp]
    \centering
    \includegraphics[width=\textwidth]{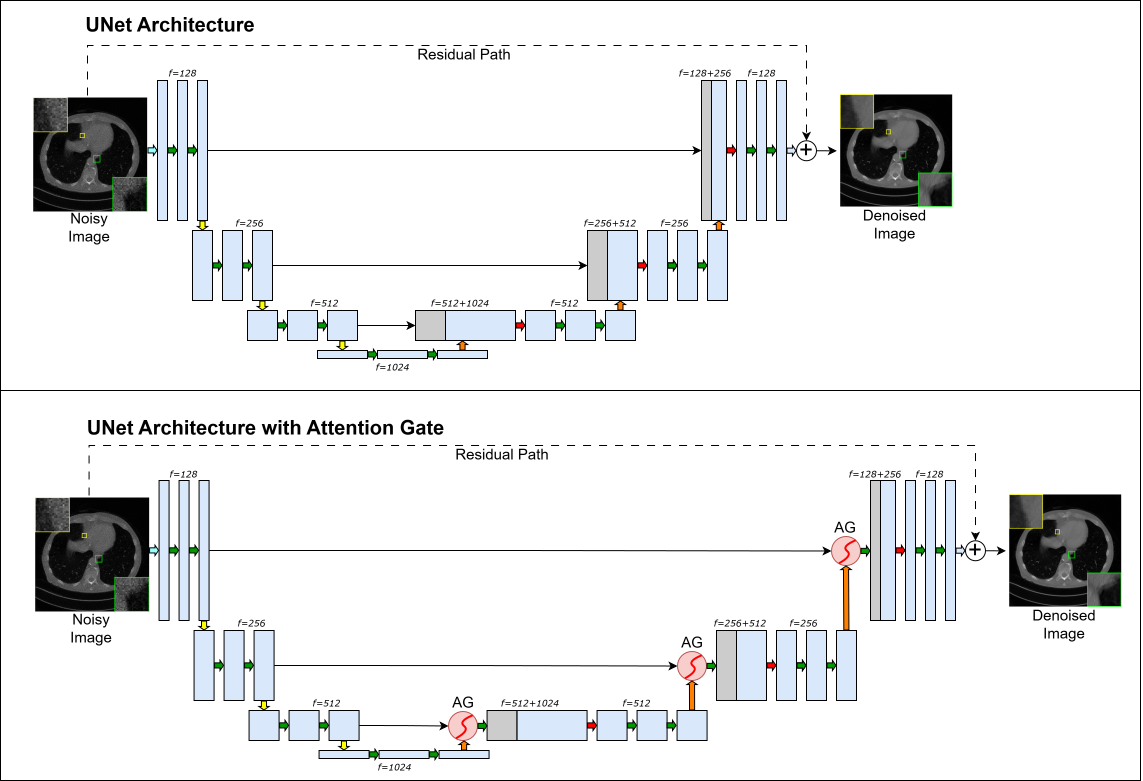}
    \caption{\textbf{Comparison of U-Net generator variants used within the CycleGAN framework for CT denoising.} 
(Top) The standard U-Net architecture augmented with a global residual path, where the network learns a residual noise component added back to the input image to produce the denoised result. 
(Bottom) The Attention U-Net variant, which incorporates attention gates (AG) along skip connections to enhance feature selection and focus on anatomically relevant regions. 
In our experiments, additional U-Net variants were also explored, including a Residual Dense U-Net (ResU-Net) that introduces local residual connections within each convolutional block.}
    \label{fig:cyclegan_generator}
\end{figure*}

The generator follows a U-Net–like encoder–decoder~\cite{Ronneberger2015UNet} with residual skip connections. As illustrated in Fig.~\ref{fig:cyclegan_generator}, the residual path allows the network to predict only the noise component, which is then added back to the input image to produce the final denoised output. 

\subsubsection{Discriminator Architecture (PatchGAN)}

\begin{figure}[H]
    \centering
    \includegraphics[width=0.99\linewidth]{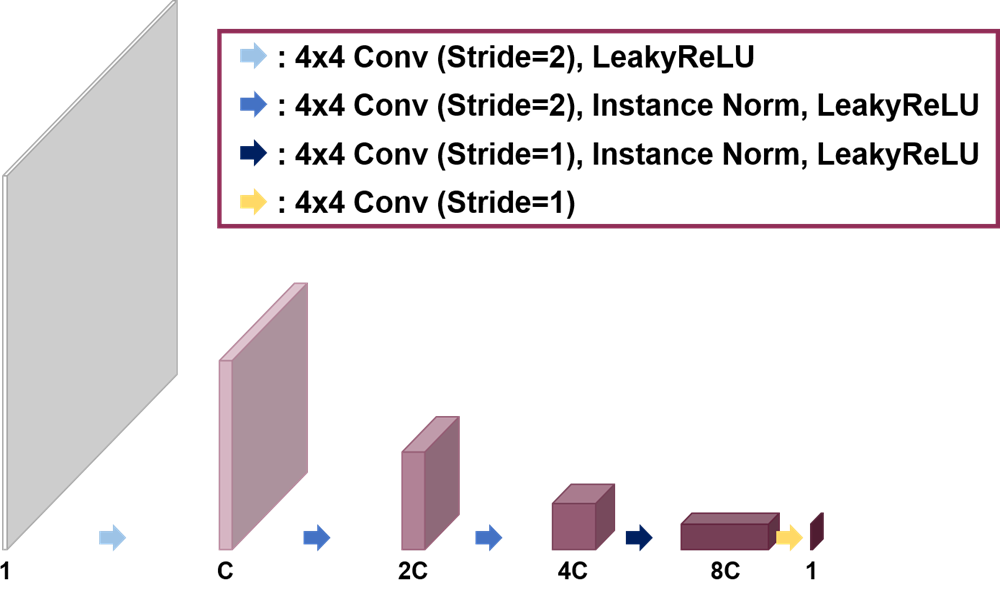}
    \caption{\textbf{PatchGAN discriminator architecture with progressive $4\times4$ convolutions.}
Each convolutional layer doubles the feature depth and reduces spatial resolution, enabling the network to classify local image patches rather than the entire image.}
    \label{fig:cyclegan_discriminator}
\end{figure}

Each discriminator adopts a PatchGAN structure~\cite{Isola2017Pix2PixPatchGAN }, classifying local image patches rather than the entire image, enabling better texture preservation. The detailed configuration is shown in Fig.~\ref{fig:cyclegan_discriminator}, consisting of four convolutional layers with $4\times4$ kernels. The channel width doubles at each layer, and LeakyReLU activations with instance normalization are applied.

\subsubsection{Loss Functions}

\begin{figure*}[!t]
    \centering
    \includegraphics[width=\textwidth]{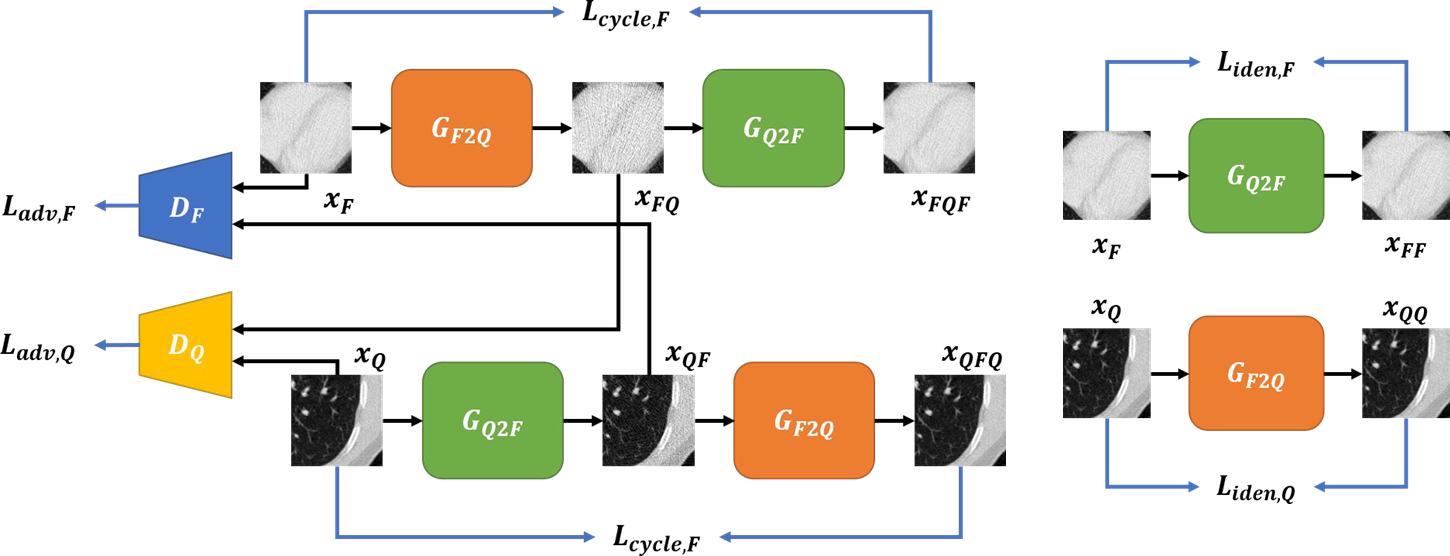}
    \caption{\textbf{Comprehensive overview of the loss functions used in the proposed CycleGAN-based residual learning framework for CT denoising.} 
    The adversarial losses ($\mathcal{L}_{adv,F}$ and $\mathcal{L}_{adv,Q}$) encourage each generator to produce realistic images that fool their respective discriminators. The cycle-consistency losses 
    ($\mathcal{L}_{cycle,F}$ and $\mathcal{L}_{cycle,Q}$) enforce that a sample translated from one domain and then back to the original 
    domain remains structurally consistent, thereby preserving anatomical fidelity. The identity losses 
    ($\mathcal{L}_{iden,F}$ and $\mathcal{L}_{iden,Q}$) constrain each generator to behave as an identity mapping when the input already 
    belongs to its target domain, preventing over-modification of high-quality inputs.}
    \label{fig:cyclegan_losses}
\end{figure*}

The CycleGAN~\cite{Zhu2017CycleGAN} training objective combines three primary losses (as shown in Fig.~\ref{fig:cyclegan_losses}): adversarial loss, cycle-consistency loss, and identity loss.

\paragraph{Adversarial Loss.}
Two discriminators, $D_F$ and $D_Q$, are used to distinguish between real and generated images in the normal-dose (F) and low-dose (Q) domains, respectively. The least-squares GAN (LSGAN) formulation is used for stability:
\begin{align}
    \mathcal{L}_{adv,F} &= 
        \mathbb{E}_{x_F}[(D_F(x_F) - 1)^2] + 
        \mathbb{E}_{x_Q}[D_F(G_{Q2F}(x_Q))^2], \\
    \mathcal{L}_{adv,Q} &=
        \mathbb{E}_{x_Q}[(D_Q(x_Q) - 1)^2] +
        \mathbb{E}_{x_F}[D_Q(G_{F2Q}(x_F))^2].
\end{align}

\paragraph{Cycle-Consistency Loss}
To maintain anatomical consistency between translations, we enforce:
\begin{align}
    \mathcal{L}_{cycle,F} &= 
        \mathbb{E}_{x_F}\left[\| G_{Q2F}(G_{F2Q}(x_F)) - x_F \|_1\right], \\
    \mathcal{L}_{cycle,Q} &= 
        \mathbb{E}_{x_Q}\left[\| G_{F2Q}(G_{Q2F}(x_Q)) - x_Q \|_1\right].
\end{align}

\paragraph{Identity Loss.}
To prevent unnecessary distortion when feeding an image already in the target domain, we use:
\begin{align}
    \mathcal{L}_{iden,F} &= 
        \mathbb{E}_{x_F}\left[\| G_{Q2F}(x_F) - x_F \|_1\right], \\
    \mathcal{L}_{iden,Q} &= 
        \mathbb{E}_{x_Q}\left[\| G_{F2Q}(x_Q) - x_Q \|_1\right].
\end{align}

\subsubsection{Training Objective and Optimization}

The full generator objective is
\begin{align}
\mathcal{L}_{G} = &\, (\mathcal{L}_{adv,F} + \mathcal{L}_{adv,Q}) \nonumber\\
&+ \lambda_{cycle}\big(\mathcal{L}_{cycle,F} + \mathcal{L}_{cycle,Q}\big) \nonumber\\
&+ \lambda_{iden}\big(\mathcal{L}_{iden,F} + \mathcal{L}_{iden,Q}\big)
\end{align}
while the discriminator objectives are
\begin{equation}
\mathcal{L}_{D_F}=\frac{1}{2}\mathcal{L}_{adv,F}, \qquad
\mathcal{L}_{D_Q}=\frac{1}{2}\mathcal{L}_{adv,Q}.
\end{equation}

\begin{algorithm}[H]
\caption{Training CycleGAN for CT Denoising}
\label{alg:train_cyclegan}
\begin{algorithmic}[1]
\STATE \textbf{procedure} \textsc{TrainCycleGAN}
\STATE Draw a minibatch $\{x_F^{(1)},\ldots,x_F^{(m)}\}\subset X_F$
\STATE Draw a minibatch $\{x_Q^{(1)},\ldots,x_Q^{(m)}\}\subset X_Q$
\STATE Compute generator loss:
\STATE \hspace{0.9em}
$\mathcal{L}_{G}=(\mathcal{L}_{adv,F}+\mathcal{L}_{adv,Q})
+\lambda_{cycle}(\mathcal{L}_{cycle,F}+\mathcal{L}_{cycle,Q})
+\lambda_{iden}(\mathcal{L}_{iden,F}+\mathcal{L}_{iden,Q})$
\STATE Update generators $G_{F2Q},\,G_{Q2F}$ by descending $\nabla_{\theta_G}\mathcal{L}_{G}$
\STATE Compute discriminator losses:
\STATE \hspace{0.9em}
$\mathcal{L}_{D_F}=\tfrac{1}{2}\mathcal{L}_{adv,F}$,\quad
$\mathcal{L}_{D_Q}=\tfrac{1}{2}\mathcal{L}_{adv,Q}$
\STATE Update discriminators $D_F,\,D_Q$ by descending $\nabla_{\theta_D}(\mathcal{L}_{D_F}+\mathcal{L}_{D_Q})$
\STATE \textbf{end procedure}
\end{algorithmic}
\end{algorithm}

This residual CycleGAN formulation ensures that denoising focuses on removing noise patterns while preserving fine structures. The use of LSGAN loss improves training stability, and identity regularization prevents over-smoothing. Together, these design choices provide a robust unpaired learning approach for CT denoising.

\subsection{Noise2Score-Based CT Denoising}
Noise2Score (N2S)~\cite{Kim2021Noise2Score} frames denoising as posterior-mean estimation via Tweedie’s formula using a score function learned from \emph{only noisy} images. We adopt an Amortized Residual Denoising Autoencoder (AR-DAE)~\cite{Lim2020ARDAE} to estimate the score, and apply the appropriate Tweedie mapping at inference according to the assumed noise distribution.

\subsubsection{Design and Flow}
\begin{figure}[H]
    \centering
    \includegraphics[width=0.99\columnwidth]{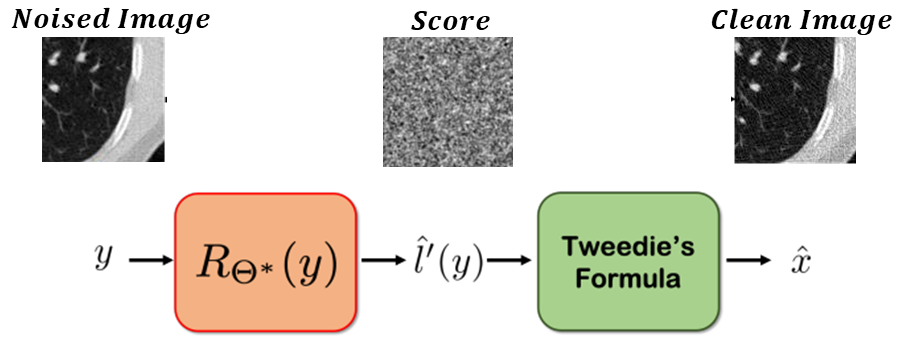}
    \caption{\textbf{Noise2Score / AR-DAE training flow: noisy image $y$ is perturbed by $u\!\sim\!\mathcal{N}(0,I)$ with scale $\sigma_a$ and fed to $R_\Theta$.} The residual output is trained to provide the score needed for Tweedie’s denoising at test time.}
    \label{fig:n2s_flow_training}
\end{figure}

Fig.~\ref{fig:n2s_flow_training} shows the AR-DAE training flow: a noisy input $y$ is perturbed by Gaussian $u\!\sim\!\mathcal{N}(0,I)$ scaled by $\sigma_a$ and passed through a residual denoiser $R_\Theta(\cdot)$ to produce a score-proportional output used in the AR-DAE loss. At inference (Fig.~\ref{fig:n2s_flow_training}), the learned score estimate $\hat{\ell}'(y)\!\approx\!\nabla_y\log p_Y(y)$ is plugged into the Tweedie estimator to recover $\hat{x}$.

\subsubsection{AR-DAE Objective}
Given noisy samples $y\!\sim\!P_Y$, Gaussian perturbations $u\!\sim\!\mathcal{N}(0,I)$, and random augmentation scale $\sigma_a\!\sim\!\mathcal{N}(0,\delta^2)$, the AR-DAE~\cite{Lim2020ARDAE} loss is
\begin{equation}
\label{eq:ardae}
\mathcal{L}_{\mathrm{AR\text{-}DAE}}(\Theta)=
\mathbb{E}_{y,u,\sigma_a}\big\|\,u+\sigma_a\,R_\Theta\big(y+\sigma_a u\big)\big\|_2^2,
\end{equation}
which trains $R_\Theta(\cdot)$ to approximate the score of $P_Y$ (up to a known scale). We use the \emph{same backbone} as our CycleGAN generator (U-Net~\cite{Ronneberger2015UNet} variants with a global residual path).

\subsubsection{Tweedie Inference}
At test time, a distribution-specific Tweedie estimator maps the noisy observation $y$ and the learned score $\hat{\ell}'(y)$ to the posterior mean $\hat{x}$. Table~\ref{tab:tweedie_cases} lists special cases used in practice.

\begin{table}[H]
\caption{Special cases of Tweedie’s formula for denoising}
\label{tab:tweedie_cases}
\centering
\begin{tabular}{lccc}
\hline
\textbf{Distribution} & $\boldsymbol{\rho}$ & $\boldsymbol{\phi}$ & $\boldsymbol{\hat{\mu}}$ \\ \hline
Gaussian & $0$ & $\sigma^{2}$ &
$\displaystyle \hat{\mu}=y+\sigma^{2}\,\hat{\ell}'(y)$ \\[3pt]
Poisson  & $1$ & $\zeta$ &
$\displaystyle \hat{\mu}=\big(y+\tfrac{\zeta}{2}\big)\exp\!\big(\zeta\,\hat{\ell}'(y)\big)$ \\[3pt]
Gamma$(\alpha,\alpha)$ & $2$ & $1/\alpha$ &
$\displaystyle \hat{\mu}=\frac{\alpha\,y}{(\alpha-1)-y\,\hat{\ell}'(y)}$ \\ \hline
\end{tabular}
\end{table}

\subsubsection{Algorithms}

The Noise2Score framework is trained using an Amortized Residual Denoising Autoencoder (AR-DAE) to estimate the score function of noisy CT images. 
\begin{algorithm}[H]
\caption{Training Noise2Score (AR-DAE)}
\label{alg:n2s_train}
\begin{algorithmic}[1]
\STATE \textbf{Input:} noisy dataset $\{y\}$, noise-aug. scale distribution $\sigma_a\sim\mathcal{N}(0,\delta^2)$
\STATE \textbf{repeat}
\STATE \quad Sample minibatch $y\sim P_Y$, $u\sim\mathcal{N}(0,I)$, $\sigma_a$
\STATE \quad $\tilde{y}\leftarrow y+\sigma_a u$ \hfill (noise augmentation)
\STATE \quad $r\leftarrow R_\Theta(\tilde{y})$ \hfill (AR-DAE forward)
\STATE \quad $\mathcal{L}\leftarrow \|\,u+\sigma_a r\,\|_2^2$ \hfill (Eq.~\eqref{eq:ardae})
\STATE \quad Update $\Theta\leftarrow\Theta-\eta\nabla_\Theta \mathcal{L}$
\STATE \textbf{until} convergence
\end{algorithmic}
\end{algorithm}

At inference time, the trained score network $R_\Theta$ predicts the score $\hat{\ell}'(y)$, which is then used in Tweedie’s formula to obtain the denoised estimate $\hat{x}$. 
\begin{algorithm}[H]
\caption{Noise2Score Inference via Tweedie’s Formula}
\label{alg:n2s_infer}
\begin{algorithmic}[1]
\STATE \textbf{Input:} noisy image $y$, learned AR-DAE $R_\Theta$, chosen noise model $(\rho,\phi)$
\STATE Compute score: $\hat{\ell}'(y)\leftarrow R_\Theta(y)$
\STATE \textbf{if} Gaussian $(\rho{=}0,\phi{=}\sigma^2)$ \textbf{then} $\hat{\mu}\leftarrow y+\sigma^{2}\hat{\ell}'(y)$
\STATE \textbf{else if} Poisson $(\rho{=}1,\phi{=}\zeta)$ \textbf{then} $\hat{\mu}\leftarrow (y+\tfrac{\zeta}{2})\exp(\zeta\hat{\ell}'(y))$
\STATE \textbf{else if} Gamma$(\alpha,\alpha)$ $(\rho{=}2,\phi{=}1/\alpha)$ \textbf{then} $\hat{\mu}\leftarrow \dfrac{\alpha\,y}{(\alpha-1)-y\,\hat{\ell}'(y)}$
\STATE \textbf{return} $\hat{x}\leftarrow \hat{\mu}$
\end{algorithmic}
\end{algorithm}

We set the AR-DAE backbone to match the CycleGAN generator (U-Net~\cite{Ronneberger2015UNet} / ResU-Net~\cite{Gurrola2021ResidualDenseUNet} / Attention U-Net~\cite{Oktay2018AttentionUNet}) with a global residual path and sample $\sigma_a$ per-batch to stabilize score learning. At test time, the correct Tweedie case must match the assumed/ injected noise (e.g., Poisson for quantum noise).

\section{Experiments and Results}
\label{sec:exp}


\subsection{Configuration and Model Selection}
\label{subsec:config}
We first performed a configuration sweep for CycleGAN across ten settings and three generator families (Standard U\,-Net~\cite{Ronneberger2015UNet}, ResU\,-Net$+$~\cite{Gurrola2021ResidualDenseUNet}, and Attention U\,-Net~\cite{Oktay2018AttentionUNet}); the full grid and results are summarized in Table~\ref{tab:cyclegan-sweep}. All models used Adam with $\beta_1{=}0.5$, $\beta_2{=}0.999$, learning rate $2\times10^{-4}$, and $100$ epochs; the estimated score was defined as
$\mathrm{est.}=\mathrm{PSNR}/40+\mathrm{SSIM}$. The best generalizing configuration on the unseen set (green row in Table~\ref{tab:cyclegan-sweep}) is the \emph{Standard U\,-Net} with $\lambda_{\text{cycle}}{=}30$, $\lambda_{\text{iden}}{=}2$, and $(ngf,ndf)=(64,64)$.

\begin{table*}[!t]
\centering
\caption{CycleGAN CT Denoising: configuration sweep and results}
\label{tab:cyclegan-sweep}
\renewcommand{\arraystretch}{1.1}
\setlength{\tabcolsep}{4pt}
\begin{threeparttable}
\begin{tabular}{c l c c c c c c c c}
\toprule
\multirow{2}{*}{\#} & \multirow{2}{*}{Architecture}
& \multicolumn{4}{c}{\textbf{CONFIGURATION}}
& \multicolumn{4}{c}{\textbf{RESULT}} \\
\cmidrule(lr){3-6}\cmidrule(lr){7-10}
 &  & $\lambda_{\text{cycle}}$ & $\lambda_{\text{iden}}$ & ngf & ndf
 & PSNR$\uparrow$ & SSIM$\uparrow$ & est. Score & ref. Score \\
\midrule
1 & Standard U\,-Net  & 10 & 5   & 64 & 64 & 37.923 & 0.959 & 1.907 & 1.892 \\
2 & Standard U\,-Net  & 15 & 7.5 & 96 & 96 & 38.128 & 0.967 & 1.920 & 1.903 \\
\rowcolor{green!12}  
3 & \textbf{Standard U\,-Net } & \textbf{30} & \textbf{2} & \textbf{64} & \textbf{64} & \textbf{38.480} & \textbf{0.967} & \textbf{1.929} & \textbf{1.923} \\
4 & Standard U\,-Net  & 25 & 2   & 64 & 48 & 38.591 & 0.964 & 1.929 & 1.912 \\
5 & ResU\,-Net+       & 10 & 5   & 64 & 64 & 38.508 & 0.968 & 1.931 & 1.917 \\
6 & ResU\,-Net+       & 30 & 2   & 64 & 64 & 38.313 & 0.963 & 1.921 & N/N \\
7 & ResU\,-Net+       & 20 & 2   & 64 & 64 & 37.585 & 0.955 & 1.895 & N/N \\
8 & Attention U\,-Net & 30 & 2   & 64 & 64 & 38.577 & 0.965 & 1.929 & 1.900 \\
9 & Attention U\,-Net & 10 & 5   & 64 & 64 & 38.254 & 0.963 & 1.919 & N/N \\
10& Attention U\,-Net & 20 & 5   & 64 & 64 & 37.924 & 0.948 & 1.896 & N/N \\
\bottomrule
\end{tabular}
\begin{tablenotes}[flushleft]
\footnotesize
\item \textbf{Remark.} All configurations use learning rate $2\times10^{-4}$ for 100 epochs with Adam $(\beta_1{=}0.5,\ \beta_2{=}0.999)$. Estimated score: $\text{est. Score}=\text{PSNR}/40+\text{SSIM}$. “ref. Score” is the score on an unseen set (Kaggle leaderboard); “N/N” = not submitted.
\end{tablenotes}
\end{threeparttable}
\end{table*}

\begin{table}[t]
\centering
\caption{Selected best CycleGAN configuration}
\label{tab:best-cyclegan}
\small
\setlength{\tabcolsep}{4pt}
\begin{threeparttable}
\begin{tabular}{l l}
\toprule
\multicolumn{2}{l}{\textbf{CONFIGURATION}}\\
\midrule
Architecture           & Standard U\,-Net \\
$\lambda_{\text{cycle}}$ & 30 \\
$\lambda_{\text{iden}}$  & 1 \\
ngf / ndf              & 64 / 64 \\
$lr_G$ / $lr_D$        & $1\times10^{-4}$ / $2\times10^{-4}$ \\
Optimizer              & Adam $(\beta_1{=}0.5,\ \beta_2{=}0.999)$ \\
Max epochs             & 1000 \\
LR scheduler           & decay at epoch 100 \\
\midrule
\multicolumn{2}{l}{\textbf{INPUT}}\\
\midrule
avg. noisy PSNR          & 34.66212 dB \\
avg. noisy SSIM          & 0.92340 \\
\midrule
\multicolumn{2}{l}{\textbf{RESULT}}\\
\midrule
avg. PSNR    & 38.91372 dB ($\uparrow$ 4.251 dB) \\
avg. SSIM    & 0.97129 ($\uparrow$ 5.186 \%) \\
est. Score   & 1.94413  \\
ref. Score   & 1.93433 \\
\bottomrule
\end{tabular}
\begin{tablenotes}[flushleft]
\footnotesize
\item \textbf{Remark.} Estimated score: $\text{PSNR}/40+\text{SSIM}$. “ref. Score” is the score on an unseen set (Kaggle leaderboard).
\end{tablenotes}
\end{threeparttable}
\end{table}

We then trained the selected CycleGAN to convergence with a longer schedule (Table~\ref{tab:best-cyclegan}): $1000$ epochs, $lr_G=1\times10^{-4}$, $lr_D=2\times10^{-4}$, and a learning-rate decay at epoch $100$. For the self-supervised baseline, the best Noise2Score hyperparameters are listed in Table~\ref{tab:best-n2s} ($\zeta=0.01$, learning rate $1\times10^{-4}$, Adam, $100$ epochs).

\subsection{Quantitative Results}
\label{subsec:quant}
The selected CycleGAN model (Table~\ref{tab:best-cyclegan}) improves the noisy input from $\text{PSNR}=34.6621$\,dB and $\text{SSIM}=0.9234$ to
$\text{PSNR}=38.9137$\,dB and $\text{SSIM}=0.9712$, i.e., a gain of
$\approx +4.251$\,dB PSNR and $+0.0478$ SSIM (about $+5.186\%$ relative).
Its estimated score is $1.94413$, and it attains a reference score of $1.93433$ on the unseen set.

\begin{table}[H]
\centering
\caption{Selected best Noise2Score configuration}
\label{tab:best-n2s}
\small
\setlength{\tabcolsep}{4pt}
\begin{tabular}{l l}
\toprule
\multicolumn{2}{l}{\textbf{CONFIGURATION}}\\
\midrule
$\zeta$                & 0.01 \\
Learning Rate          & $4\times10^{-4}$ \\
Optimizer              & Adam $(\beta_1{=}0.5,\ \beta_2{=}0.999)$ \\
Epochs                 & 100 \\
\midrule
\multicolumn{2}{l}{\textbf{INPUT}}\\
\midrule
avg. noisy PSNR          & 25.57495 dB \\
avg. noisy SSIM          & 0.73098 \\
\midrule
\multicolumn{2}{l}{\textbf{RESULT}}\\
\midrule
avg. PSNR    & 34.27972 dB ($\uparrow$ 8.704 dB) \\
avg. SSIM    & 0.92564 ($\uparrow$ 26.63 \%) \\
\bottomrule
\end{tabular}
\end{table}

Noise2Score (Table~\ref{tab:best-n2s}) shows a strong improvement from a much noisier starting point:
$\text{PSNR}=25.57495$\,dB and $\text{SSIM}=0.73098$ to
$\text{PSNR}=34.27972$\,dB and $\text{SSIM}=0.92564$ (i.e., $+8.704$\,dB PSNR and $+0.1946$ SSIM, roughly $+26.63\%$ relative).
Although the absolute \emph{final} scores favor CycleGAN on our data (higher PSNR/SSIM), Noise2Score recovers a larger margin over its noisy input, reflecting its effectiveness without paired supervision.

\subsection{Qualitative Comparison}
\label{subsec:qual}

\begin{figure*}[htbp]
  \centering
  \includegraphics[width=0.86\textwidth]{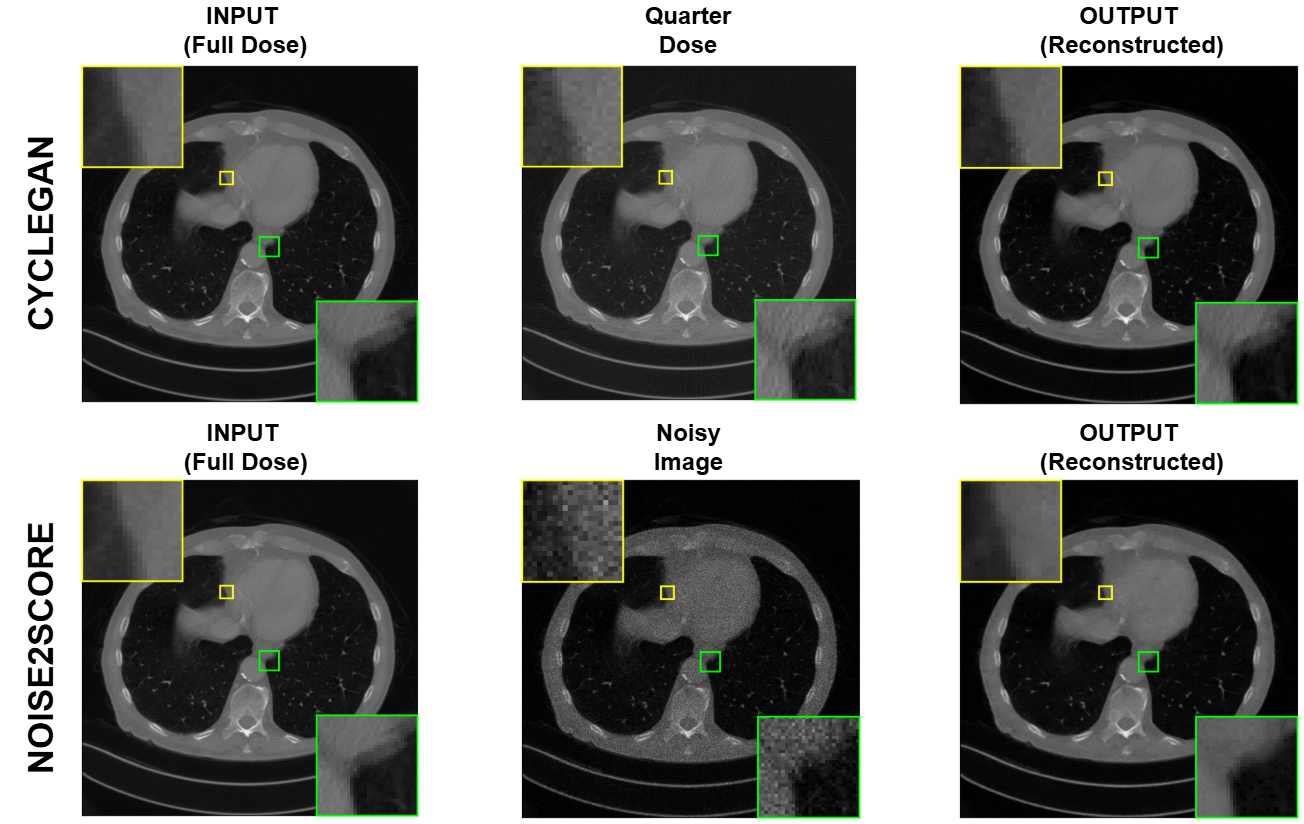}
  \caption{\textbf{CycleGAN CT denoising: qualitative comparison on a representative axial slice.} Yellow and green insets mark magnified ROIs to show noise suppression.}
  \label{fig:unet-qual}
\end{figure*}

Figure~\ref{fig:unet-qual} compares visual outcomes for both methods on a representative axial slice. Both approaches suppress quantum noise and streaking artifacts, but CycleGAN tends to preserve edges and soft-tissue boundaries more sharply in the magnified ROIs, whereas Noise2Score occasionally exhibits mild texture smoothing or residual speckle in challenging regions. In particular, edges near bone–soft-tissue interfaces and vessel boundaries appear crisper in the CycleGAN reconstructions, while remaining free of over-enhancement artifacts.

\subsection{Training Dynamics}
\label{subsec:dynamics}

\begin{figure*}[!t]
  \centering
  \begin{subfigure}[t]{0.49\linewidth}
    \centering
    \includegraphics[width=\linewidth]{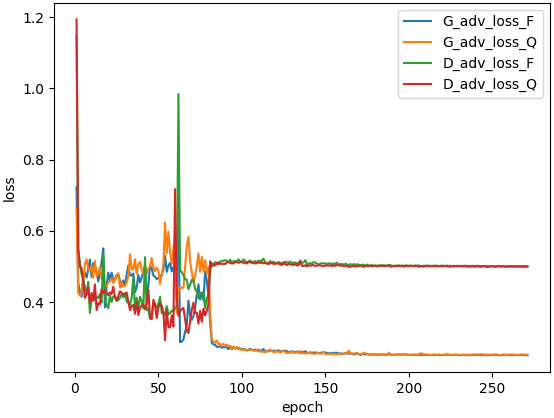}
    \caption{Adversarial losses}
  \end{subfigure}\hfill
  \begin{subfigure}[t]{0.509\linewidth}
    \centering
    \includegraphics[width=\linewidth]{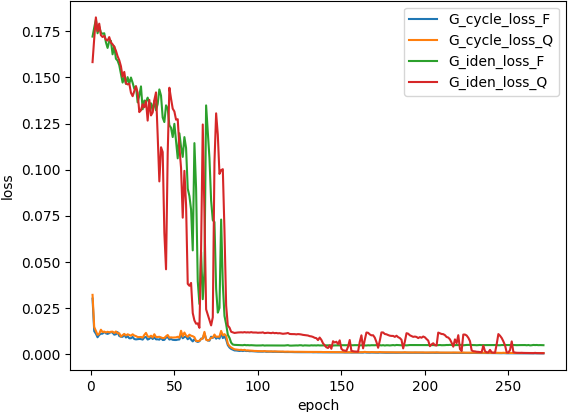}
    \caption{Cycle \& identity losses}
  \end{subfigure}
  \caption{\textbf{Training loss curves in CycleGAN.} After early transients, losses stabilize following the learning-rate schedule change, with occasional pre-decay spikes (more pronounced for the $Q$-identity loss).}
  \label{fig:cyclegan-losses}
\end{figure*}



\begin{figure}[H]
    \centering
    \includegraphics[width=0.89\columnwidth]{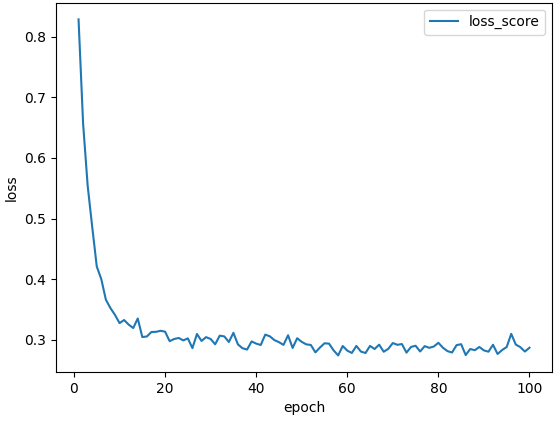}
    \caption{\textbf{Training loss curves in the Noise2Score.}}
    \label{fig:n2s_loss_curves}
\end{figure}
The training curves of CycleGAN are shown in Fig.~\ref{fig:cyclegan-losses}. After an initial transient, the adversarial losses oscillate and then undergo a clear transition near the learning-rate schedule change, after which the generator adversarial losses settle to a lower baseline ($\sim0.25$) while the discriminator losses hover near~0.5. The cycle-consistency losses drop sharply around the same point and remain close to zero thereafter. The identity losses also decay; the $Q$-domain identity loss exhibits larger pre-decay fluctuations and occasional spikes, which largely disappear once the schedule takes effect. On the other hand, the Noise2Score objective (Fig.~\ref{fig:n2s_loss_curves}) decreases rapidly within the first $\sim10$ epochs and then plateaus with small fluctuations, indicating stable convergence under the chosen $\zeta$ and learning rate.

\subsection{Discussion}
\label{subsec:discussion}
From the ablation in Table~\ref{tab:cyclegan-sweep}, the Standard U\,-Net with a stronger cycle term ($\lambda_{\text{cycle}}{=}30$) and weaker identity term ($\lambda_{\text{iden}}{=}2$) consistently balances data fidelity and realism; larger identity weights tended to over-smooth high-frequency details, while attention variants did not surpass the simpler backbone under the same budget. Overall, the selected CycleGAN achieves the best absolute PSNR/SSIM and generalization to the unseen set, whereas Noise2Score delivers competitive denoising without paired targets and achieves large relative gains from very noisy inputs. These complementary strengths suggest a practical strategy: use Noise2Score when paired data are unavailable, and prefer CycleGAN when adversarial training can leverage domain transfer with cycle/identity regularization to maximize final image quality.

\section{Conclusion}
We compared a CycleGAN-based residual translator and a Noise2Score denoiser for CT image denoising under consistent protocols. The CycleGAN configuration selected via a systematic sweep produced the highest absolute PSNR/SSIM and the strongest generalization to an unseen set (Kaggle), while Noise2Score delivered substantial relative improvements from very noisy inputs without any clean targets. These findings suggest a pragmatic guideline: when unpaired domain translation is feasible, CycleGAN with strong cycle and light identity regularization maximizes final image quality; when only noisy data are available, Noise2Score provides a competitive, pair-free alternative with stable training and robust gains.

\bibliographystyle{IEEEtran}
\bibliography{myblib}
\end{document}